%% file: main.tex
\documentclass[10pt,twocolumn,letterpaper]{article}

\usepackage{cvpr}              

\input{preamble}

\definecolor{cvprblue}{rgb}{0.21,0.49,0.74}
\usepackage[pagebackref,breaklinks,colorlinks,allcolors=cvprblue]{hyperref}
\usepackage{multirow}

\usepackage{algorithm}
\usepackage[noend]{algpseudocode}

\def\confName{CVPR}
\def\confYear{2026}

\title{BitTP: The Lightweight Trajectory Prediction Model\\
with BitLLM for Edge-Devices}

\author{
Mincheol Kang\textsuperscript{1,*,$\ddagger$} \quad
Hyunjin Lim\textsuperscript{2,*} \quad
Bomin Kang\textsuperscript{2,*} \quad
Daehee Park\textsuperscript{2,$\dagger$} \\
\textsuperscript{1}KAIST, Republic of Korea \textsuperscript{2}DGIST, Republic of Korea \\
{\tt mintcat@kaist.ac.kr}, \quad
{\tt \{hyunjinlim, mellowoura, dhpark\}@dgist.ac.kr}
}

\begin{document}
\maketitle

\renewcommand{\thefootnote}{\fnsymbol{footnote}} 
\footnotetext[1]{Equal contribution.}
\footnotetext[2]{Corresponding author.}
\footnotetext[3]{ Work done during an undergraduate internship at DGIST.}
\renewcommand{\thefootnote}{\arabic{footnote}} 

\input{sec/0_abstract}
\input{sec/1_intro}
\input{sec/2_rws}

\input{sec/3_method}
\input{sec/4_exps}
\input{sec/5_conclusion}
\newpage
\input{sec/6_acknowledgement}
{
    \small
    \bibliographystyle{ieeenat_fullname}
    \bibliography{main}
}

\setcounter{section}{0} 
\renewcommand{\thesection}{S\arabic{section}} 
\setcounter{figure}{0}
\renewcommand{\thefigure}{S\arabic{figure}}
\setcounter{table}{0}
\renewcommand{\thetable}{S\arabic{table}}
\setcounter{equation}{0}
\renewcommand{\theequation}{S\arabic{equation}}

\input{sec/X_suppl}

\end{document}

%% file: preamble.tex
\usepackage[dvipsnames]{xcolor}
\usepackage{array}
\newcolumntype{C}{>{\centering\arraybackslash}m{2.0cm}}

\usepackage{array}
\usepackage{multirow}

%% file: sec/0_abstract.tex
\begin{abstract}

Trajectory prediction is a fundamental task for autonomous systems, requiring complex reasoning about multi-agent interactions and intents.
Large language models (LLMs) have recently been adopted for this task, as they provide strong contextual reasoning and interpretable, language-based trajectory representations.
However, these LLM-based predictors are extremely memory- and compute-intensive, making them difficult to deploy on resource-constrained edge devices such as on-board computers in autonomous robots.
To bridge this gap, we propose BitTP, which converts an LLM-based trajectory predictor into a lightweight bitlinear architecture. We demonstrate that weight-only quantization to 1.58-bit (BitTP-Weight) is optimal. Crucially, activations must remain in full precision, as quantizing them leads to severe degradation and instability in spatio-temporal reasoning.
Empirically, BitTP-Weight not only preserves but improves prediction quality over the full-precision (BF16) LLM baseline, reducing ADE by 14.29\% and FDE by 20.97\% on average, while simultaneously reducing memory usage and inference latency relative to other quantization methods. These results demonstrate that carefully designed quantization acts as an effective regularizer, enabling the practical deployment of sophisticated LLM-based reasoning on edge devices. Code is available at: \url{https://github.com/MintCat98/BitTP}.
\end{abstract}

%% file: sec/1_intro.tex
\section{Introduction}
\label{sec:intro}

\begin{figure}[h]
    \centering
    \includegraphics[width=1.0\linewidth]{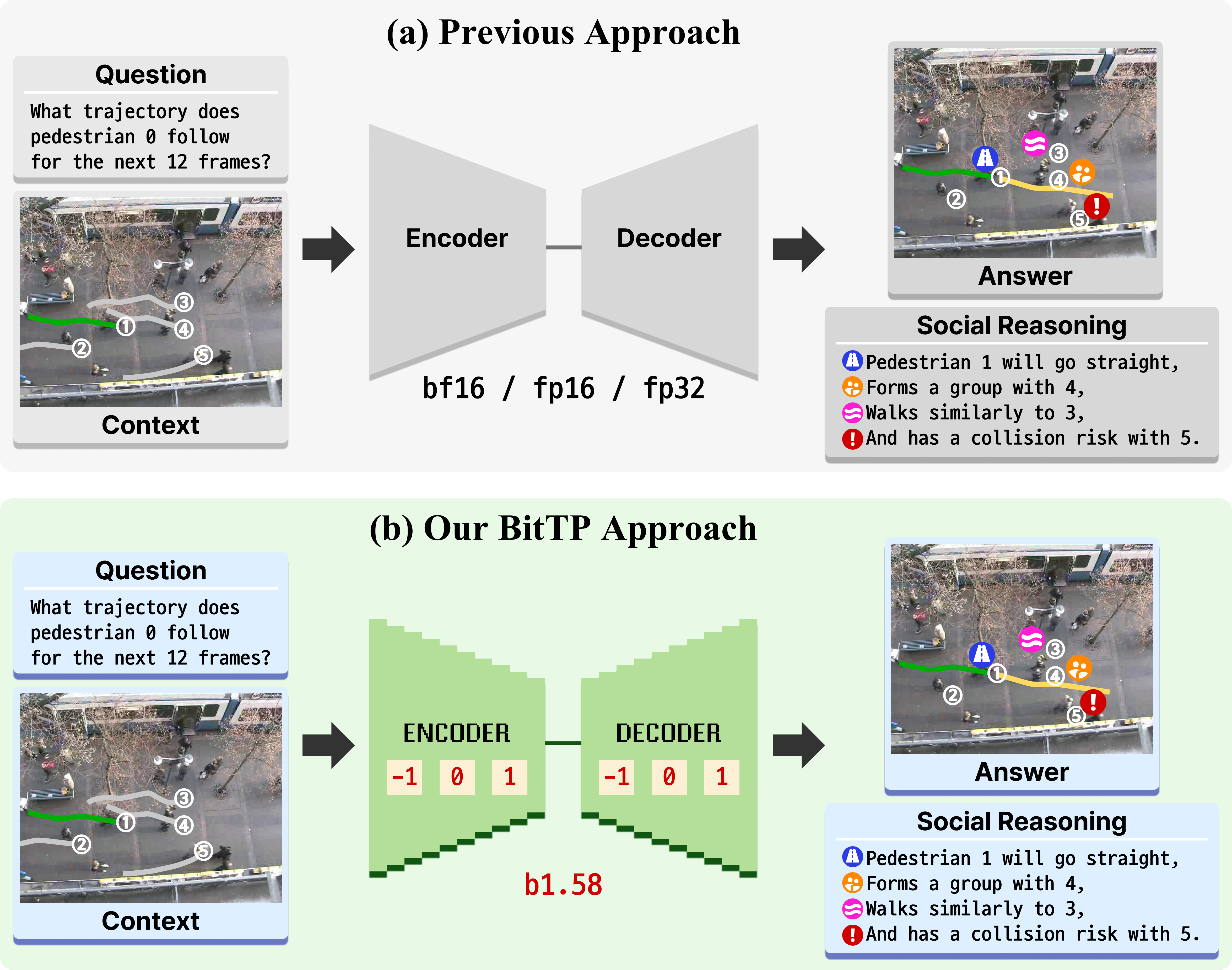}
    \caption{\textbf{Overview of the BitTP framework.} (a) The baseline trajectory prediction model \cite{Bae_2024_CVPR} utilizing a full-precision \texttt{(bf16,fp16,fp32)} sequence-to-sequence architecture. (b) Our proposed BitTP, which replaces standard \texttt{nn.Linear} layers with \texttt{BitLinear} modules to selectively quantize model weights to 1.58-bit \texttt{[-1,0,1]}.}
    \label{fig:overview}
\end{figure}

Trajectory prediction plays a pivotal role in autonomous driving and robotic navigation, acting as the bridge between perception and planning~\cite{mohamed2020socialstgcnn,huang2019stgat,yuan2021agentformer,ngiam2022scenetransformerunifiedarchitecture,zhao2020tnt,zhou2023query}.
Accurate trajectory prediction allows an autonomous agent to anticipate how surrounding agents will move in dynamic scenes and to plan safe, socially compliant actions.
However, real-world environments are highly interactive: pedestrians, vehicles, and robots influence one another, and forecasting future motions therefore requires understanding the intent and interactions of multiple agents rather than purely extrapolating past positions~\cite{alahi2016social,gupta2018social,lee2017desire,salzmann2020trajectronpp,gao2020vectornet,liang2020lanegcn}.

With this motivation, recent studies have begun to leverage the contextual reasoning abilities of large language models (LLMs) for trajectory prediction.
Language models offer strong capabilities in contextual reasoning and multi-agent understanding~\cite{brown2020gpt3,devlin2019bert,touvron2023llama,sanh2019distilbert,lan2019albert}.
Vision–language frameworks and language-driven predictors have explored encoding trajectories, intents, and scene descriptions in a language-like format to improve interpretability and scene-level consistency~\cite{zhou2025uni4d,sima2024drivelm,zhou2023unifiedframeworkmultimodalmultipart,lan2024trajllmnewexplorationempowering}.
However, the large size and heavy computational requirements of LLMs make them unsuitable for deployment on edge devices, where real-time inference and energy efficiency are essential.
This limitation prevents their direct use in autonomous vehicles and mobile robots operating under strict resource constraints.
To mitigate these constraints, many works explore parameter-efficient tuning and quantization for LLMs~\cite{hu2021lora,dettmers2023qloraefficientfinetuningquantized,frantar2023gptqaccurateposttrainingquantization,wang2023bitnet,lin2024awqactivationawareweightquantization,dettmers2022llmint8_neurips}.

In trajectory prediction, sequence-to-sequence (Seq2Seq) encoder-decoder architectures are widely recognized to outperform decoder-only models, as demonstrated by recent state-of-the-art approaches \cite{Bae_2024_CVPR}.
However, the majority of existing quantization research has concentrated on decoder-only models, particularly within the LLaMA family. Specifically, these works demonstrate that weight-only quantization effectively preserves the performance of LLaMA-like models, whereas quantizing activations often leads to significant degradation \cite{frantar2023gptqaccurateposttrainingquantization,lin2024awqactivationawareweightquantization}. It remains unverified whether this finding holds true for the encoder-decoder structure of Seq2Seq models. Our work addresses this gap by investigating efficient quantization strategies specifically for Seq2Seq architectures. We select BitNet \cite{wang2023bitnet} due to its ability to perform extreme model compression down to 1.58-bits. We hypothesize that BitNet will be particularly impactful for Seq2Seq models, as its aggressive compression can be applied to both the encoder and decoder, offering substantial size reductions ideal for edge device deployment. Furthermore, we aim to empirically validate whether a selective quantization strategy—separating weights and activations—is also the optimal approach for Seq2Seq models, drawing parallels from the observations in LLaMA-based research.

Building upon this hypothesis, we propose BitTP (Fig.~\ref{fig:overview}), a quantized reasoning framework for efficient trajectory prediction based on Seq2Seq LLMs.
Instead of relying on massive LLMs, we transform a standard language‑based prediction model into a lightweight bitlinear architecture that performs low‑bit operations while preserving reasoning capability.
Our approach applies quantization selectively: only the model weights are quantized and activations remain in full precision.
This selective design draws inspiration from extreme network quantization and sparsity studies~\cite{frankle2019lottery,rastegari2016xnor}.
Through systematic experiments, we find that selective quantization achieves the best trade-off between accuracy and efficiency, while full quantization of both weights and activations severely degrades motion reasoning quality and leads to unstable training behaviors.
The proposed design, which builds on established multi-agent reasoning and dynamic interaction models~\cite{gomez2023improving,li2020evolvegraph,li2021spatiotemporalgraphdualattentionnetwork,chen2018intentnet}, enables the model to process complete past trajectories and generate coherent multi‑step predictions, maintaining the relational reasoning and interpretability of language‑based approaches while achieving substantial improvements.
Our model is evaluated on real‑world trajectory prediction datasets, showing that it maintains prediction performance while being suitable for deployment on resource‑limited robots~\cite{caesar2020nuscenes,chang2019argoverse,ettinger2021waymo,caesar2021nuplan,houston2020one,robicquet2016sdd}.

Our main contributions are summarized as follows:
\begin{itemize}
    \item We propose BitTP, a quantized reasoning framework for edge devices, and conduct a systematic analysis of three selective quantization strategies (weight-only, activation-only, and combined) within a sequence-to-sequence architecture.
    \item We find that the BitTP-Weight strategy, which quantizes only model weights to 1.58-bit while maintaining full-precision activations, achieves the optimal trade-off. It outperforms the full-precision baseline, reducing ADE by 14.29\% and FDE by 20.97\%, while significantly reducing memory and latency relative to previous approaches.
	\item We confirm that activation precision is critical for spatio-temporal reasoning in encoder-decoder models, as strategies involving activation quantization resulted in severe instability and performance degradation.
    Our work validates the practical feasibility of deploying lightweight, accurate language-based models for real-time trajectory prediction.
\end{itemize}

%% file: sec/2_rws.tex
\section{Related Works}
\label{sec:formatting}

\subsection{Trajectory Prediction in Dynamic Environments}
Trajectory prediction has long been a core problem in autonomous driving and robotics.
Classical approaches focus on modeling the geometric evolution of agent trajectories from past motion histories.
These methods learn spatial and temporal correlations through recurrent or transformer‑based encoders and often rely on pairwise distance or attention mechanisms to capture agent interactions.
Examples include pooling‑based recurrent models, generative adversarial frameworks, graph‑based approaches and transformer architectures~\cite{alahi2016social,gupta2018social,lee2017desire,mangalam2020journeydestinationendpointconditioned,chai2019multipath,mohamed2020socialstgcnn,huang2019stgat,salzmann2020trajectronpp,gao2020vectornet,liang2020lanegcn,yuan2021agentformer,ngiam2022scenetransformerunifiedarchitecture,zhao2020tnt,zhou2023query,zhou2022hivt}.
While effective for short‑term motion forecasting, such geometry‑centric designs struggle to capture higher‑level semantics such as human intent, social norms and multi‑agent coordination\cite{zhou2023query}.
As urban scenes become increasingly complex, accurate forecasting requires reasoning not only about motion geometry but also about latent intentions and inter‑agent dependencies across the scene.

\subsection{Reasoning-based and Language-driven Prediction}
Recent research has explored reasoning‑oriented frameworks to enhance the semantic understanding of dynamic scenes.
Graph‑based relational reasoning models treat agents and their interactions as nodes and edges in evolving graphs, enabling context‑aware predictions~\cite{li2020evolvegraph,mangalam2021ynet,kosaraju2020bigat,ivanovic2019trajectron,li2021spatiotemporalgraphdualattentionnetwork,chen2018intentnet}.
Transformers have been adapted to jointly encode social interactions and temporal dynamics for multi‑agent prediction~\cite{yuan2021agentformer,ngiam2022scenetransformerunifiedarchitecture, gomez2023improving}.
Inspired by the expressive reasoning capability of language models, several works represent trajectories, intents and scene descriptions in a language‑like format to improve interpretability and scene‑level consistency~\cite{brown2020gpt3,devlin2019bert,touvron2023llama,sanh2019distilbert,lan2019albert,zhou2025uni4d,sima2024drivelm,zhou2023unifiedframeworkmultimodalmultipart,lan2024trajllmnewexplorationempowering}.
However, the computational footprint of these models remains prohibitively large.
Even compact LLMs require multi‑gigabyte memory and tera‑scale FLOPs for autoregressive inference, leading to high latency and power consumption.
This limitation makes direct deployment of language‑driven predictors impractical for real‑time autonomous systems, motivating research toward more efficient reasoning architectures.

\subsection{Efficient and Quantized Reasoning for Edge Deployment}
To enable real‑time prediction under limited compute budgets, recent works have investigated model compression, pruning and quantization techniques.
Early studies on binary neural networks and distillation paved the way for aggressive quantization and sparsity without catastrophic performance drops~\cite{hinton2015distilling,frankle2019lottery,rastegari2016xnor}.
Lightweight language models such as DistilBERT and ALBERT demonstrate that substantial parameter reductions are possible through knowledge distillation and factorized embeddings~\cite{sanh2019distilbert,lan2019albert}.
Low‑rank adaptation and parameter‑efficient fine‑tuning methods like LoRA further reduce training and inference costs~\cite{hu2021lora}.
Recent advances in quantization, including QLoRA, GPTQ, BitNet, AWQ and 8‑bit matrix multiplication, have shown that large language models can be quantized to low precision while retaining accuracy~\cite{dettmers2023qloraefficientfinetuningquantized,frantar2023gptqaccurateposttrainingquantization,wang2023bitnet,lin2024awqactivationawareweightquantization,dettmers2022llmint8_neurips}.
Nevertheless, most of these approaches target general NLP tasks or vision transformers and do not directly address the relational reasoning needed for multi‑agent trajectory prediction.
In contrast, efficient trajectory reasoning demands architectures that not only compress parameters but also retain the ability to infer interactions and intent over time.
This gap motivates our work: we propose a quantized reasoning framework for trajectory prediction that achieves strong reasoning capability with lightweight computation, enabling deployment on edge robots and autonomous vehicles.

%% file: sec/3_method.tex
\section{Method}
\begin{figure*}[ht]
    \centering
    \includegraphics[width=1.0\linewidth]{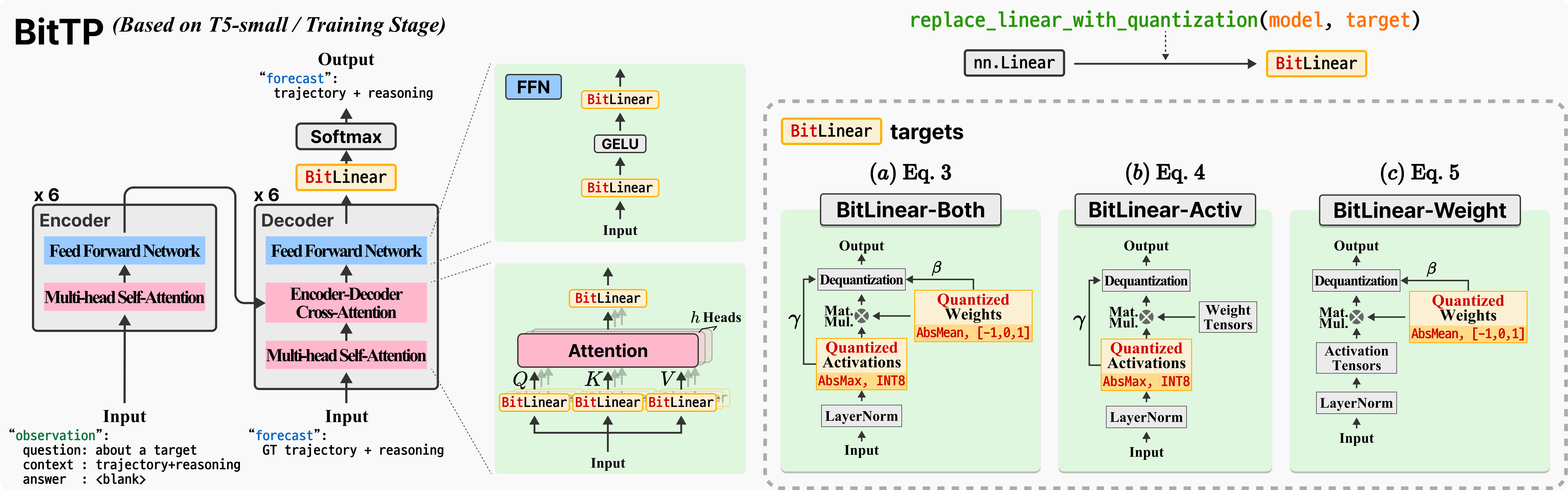}
    \caption{\textbf{BitTP Model Architecture and Selective Quantization Strategies.} \textit{(Left)} The overall BitTP framework, based on a T5-small architecture. All standard \texttt{nn.Linear} layers within the Feed-Forward Networks (FFN) and Attention blocks are dynamically replaced by our custom \texttt{BitLinear} modules. \textit{(Right)} We systematically investigate three selective quantization strategies: (a) BitLinear-Both (Eq.~\ref{eq:both}), which quantizes both 8-bit activations and 1.58-bit weights; (b) BitLinear-Activ (Eq.~\ref{eq:activ}), which quantizes only activations; and (c) BitLinear-Weight (Eq.~\ref{eq:weight}), which quantizes only weights. Our best-performing model, BitTP-Weight (c), demonstrates the effectiveness of preserving full-precision activations while compressing weights.}
    \label{fig:method}
\end{figure*}

This section elaborates on BitTP, our proposed lightweight quantization framework designed for efficient trajectory prediction on edge devices.
Sec.~\ref{sec:model_quant} introduces the backbone architecture and detailing the quantization mechanism (Alg.~\ref{alg:pseudo}) used to replace standard \texttt{nn.Linear} layers with our custom \texttt{BitLinear} modules.
Subsequently, Sec.~\ref{sec:three_strategies} presents our core technical contribution: three selective quantization strategies, namely BitLinear-Both, BitLinear-Activ, and BitLinear-Weight.
We describe the technical formulations and mathematical definitions (Eq.~\ref{eq:both}-\ref{eq:weight}) for each strategy.

\begin{algorithm}[t]
\caption{nn.Linear-to-BitLinear Replacement}
\label{alg:pseudo}
\begin{algorithmic}[1]
\Procedure{ReplaceLinearWithQuantization}{}
    \State \textbf{Input:} model $\mathcal{M}$, target $T$
    \If{$T$ \textbf{includes} Activation}
        \State activation measure $m_a \gets$ \texttt{AbsMax} \Comment{Eq.~\ref{eq:absmax}}
    \EndIf
    \If{$T$ \textbf{includes} Weight}
        \State weight measure $m_w \gets$ \texttt{AbsMean} \Comment{Eq.~\ref{eq:absmean}}
    \EndIf
    \State \Call{ReplaceLinearBase}{$\mathcal{M}$, $m_w$, $m_a$}
    \State \Return $\mathcal{M}$
\EndProcedure
\Statex
\Procedure{ReplaceLinearBase}{}
    \State \textbf{Input:} $\mathcal{M}$, $m_w$, $m_a$
    \For{$(\text{name } n, \text{module } L) \in \text{children}(\mathcal{M})$}
        \If{$L \in \mathtt{nn.Linear}$}
            \State $B \gets \mathtt{BitLinear}\!\big(\mathcal{M}, m_w, m_a\big)$ \Comment{Eq.~\ref{eq:both},\ref{eq:activ},\ref{eq:weight}}
            \State $\theta_{B} \gets \theta_{L}$
            \State \texttt{setattr}$(\mathcal{M},\,n,\, B)$
        \Else
            \State \Call{ReplaceLinearBase}{L, $m_w$, $m_a$}
        \EndIf
    \EndFor
\EndProcedure
\end{algorithmic}
\end{algorithm}

\begin{table}[t]
    \centering
    \caption{\textbf{Breakdown of nn.Linear layers in the T5-small model, which serve as the targets for quantization.}}
    \label{tab:num_linears}
    \begin{tabular}{l|ccc} 
    \hline
    \textbf{T5-small} & \textbf{Block} & \textbf{Linear (d\_ff=2048)} & \textbf{Total} \\ 
    \hline \hline 
    Encoder & 6 & \textbf{6} & 36 \\ 
    Decoder & 6 & \textbf{10} & 60 \\ 
    LM Head & - & \textbf{1} & 1 \\ 
    \hline
    \textbf{Total} & - & - & \textbf{97} \\ 
    \hline
    \end{tabular}
\end{table}

\subsection{Model Quantization}
\label{sec:model_quant}
Our proposed quantized reasoning framework, BitTP, is a general method designed to quantize standard \texttt{nn.Linear} layers, making it broadly applicable to Transformer-based models.
While language-based architectures show strong potential for trajectory prediction, recent work suggests that sequence-to-sequence (encoder–decoder) models are better suited for this task than causal language models (CLMs)~\cite{Bae_2024_CVPR}.
We therefore empirically validate our approach using the T5-small model~\cite{raffel2020exploring} as a representative encoder–decoder Transformer backbone.
BitTP adapts this backbone by replacing all \texttt{nn.Linear} layers within the T5 architecture with our custom \texttt{BitLinear} modules, as described below and in Sec.~\ref{sec:three_strategies}.

\textbf{Preliminaries. }
The T5-small model is composed of 6 encoder blocks and 6 decoder blocks, as described in Fig.~\ref{fig:method} and Tab.~\ref{tab:num_linears}.
Within each block, the standard Transformer components are our quantization targets: each Feed-Forward Network (FFN) module contains 2 \texttt{nn.Linear} layers for the gated activation, and each attention module (self-attention or cross-attention) contains 4 \texttt{nn.Linear} layers for the Q, K, V, and O projections.
As summarized in Tab.~\ref{tab:num_linears}, this architecture logically comprises 97 \texttt{nn.Linear} layers in total: 36 in the encoder, 60 in the decoder, and 1 in the final output projection layer (LM Head), which is tied with the input token embeddings (\textit{shared embedding projection}).

\textbf{Quantization Mechanism. }
To perform adaptation, we utilize a replacement helper function, \texttt{replace\_linear\_with\_quantization}.
As shown in Algorithm~\ref{alg:pseudo}, this main entry function selects the appropriate quantization measures based on the \texttt{target} string, an argument that specifies whether to quantize weights, activations, or both.
It then invokes the recursive helper function, \texttt{\_replace\_linear\_base} (Algorithm~\ref{alg:pseudo}), to traverse the T5 model's module hierarchy and replace every \texttt{nn.Linear} instance with a \texttt{BitLinear} module.
While the logical count of \texttt{nn.Linear} layers is 97 (Tab.~\ref{tab:num_linears}), we note that due to parameter sharing within the Hugging Face \texttt{transformers} library implementation \textit{(e.g., combining query, key, and value projections into a single \texttt{nn.Linear} module)}, the actual number of unique \texttt{nn.Linear} objects replaced by our function is 73.
This dynamic replacement allows us to use the same training and evaluation pipeline (Sec.~\ref{sec:experiments}) to systematically compare the performance of each quantization strategy against the full-precision baseline.

\newcommand{\negc}[1]{\textcolor{RoyalBlue}{#1}}
\newcommand{\posc}[1]{\textcolor{BrickRed}{#1}}

\begin{table*}[t]
\centering
\captionsetup{skip=4pt}
\caption{\textbf{Comparison of BitTP variants with prior state-of-the-art methods on the ETH/UCY benchmark.} All values report ADE/FDE (meters).
The $\Delta$ column shows performance change relative to the reproduced BF16 baseline (LMTraj-SUP$\dagger\dagger$), where negative values indicate improvement. BitTP results use batch size = 128.}

$\dagger$: Results reported in the original papers \textit{(may use different batch sizes)},
$\dagger\dagger$: Results from our backbone implementation.
\label{tab:bitlinear_comparison}

\resizebox{\textwidth}{!}{
\begin{tabular}{l @{\hspace{30pt}} C C C C C @{\hspace{25pt}} C C}
\toprule
\textbf{Stochastic} 
& \textbf{ETH} 
& \textbf{HOTEL} 
& \textbf{UNIV} 
& \textbf{ZARA1} 
& \textbf{ZARA2} 
& \textbf{AVG} 
& \textbf{$\Delta$} \\
\midrule

PECNet\cite{mangalam2020journeydestinationendpointconditioned}† 
& 0.61 / 1.07 & 0.22 / 0.39 & 0.34 / 0.56 & 0.25 / 0.45 & 0.19 / 0.33 & 0.32 / 0.56 & - \\

MID\cite{Gu_2022_CVPR}$\dagger$
& 0.57 / 0.93 & 0.21 / 0.33 & 0.29 / 0.55 & 0.28 / 0.50 & 0.20 / 0.37 & 0.31 / 0.54 & - \\

SocialVAE\cite{SocialVAE}$\dagger$
& 0.41 / 0.58 & 0.13 / 0.19 & 0.21 / 0.36 & 0.17 / 0.29 & 0.13 / 0.22 & 0.21 / 0.33 & - \\

LMTraj-SUP\cite{Bae_2024_CVPR}$\dagger$
& 0.41 / 0.51 & 0.12 / 0.16 & 0.22 / 0.34 & 0.20 / 0.32 & 0.17 / 0.27 & 0.22 / 0.32 & - \\
\midrule

LMTraj-SUP\cite{Bae_2024_CVPR}$\dagger\dagger$
& 0.56 / 0.82 & 0.19 / 0.35 & 0.49 / 0.98 & 0.26 / 0.46 & 0.26 / 0.50 & 0.35 / 0.62 & 0.0 / 0.0 \\

LMTraj-SUP-int8 
& 0.47 / 0.67 & 0.19 / 0.35 & 0.49 / 0.98 & 0.26 / 0.46 & 0.26 / 0.49 & 0.34 / 0.59 & \negc{-0.01} / \negc{-0.03} \\

LMTraj-SUP-int4
& 0.47 / 0.71 & 0.20 / 0.36 & 0.49 / 0.97 & 0.25 / 0.44 & 0.28 / 0.54 & 0.34 / 0.60 & \negc{-0.01} / \posc{+0.02} \\
\midrule

\textbf{BitTP-Both}
& 1.73 / 1.41 & 0.84 / 0.78 & 1.13 / 1.11 & 1.66 / 1.44 & 0.97 / 0.94 & 1.27 / 1.13 & \posc{+0.92} / \posc{+0.51} \\

\textbf{BitTP-Activation}
& 2.89 / 4.86 & 1.16 / 2.02 & 1.36 / 2.47 & 2.40 / 3.67 & 1.40 / 2.55 & 1.84 / 3.12 & \posc{+1.49} / \posc{+2.50} \\

\textbf{BitTP-Weight}
& 0.46 / 0.62 & 0.17 / 0.27 & 0.42 / 0.80 & 0.23 / 0.40 & 0.22 / 0.39 & 0.30 / 0.49 & \textbf{\negc{-0.05}} / \textbf{\negc{-0.13}} \\
\bottomrule

\end{tabular}
}
\end{table*}

\subsection{Three Selective BitLinear Strategies}
\label{sec:three_strategies}
Our primary objective is to adapt the T5-small backbone for efficient deployment on resource-constrained edge devices.
To this end, we replace the standard full-precision \texttt{nn.Linear} layers, which perform the operation $y = x W^T + b$, with custom \texttt{BitLinear} layers.
This adaptation inherits the core quantization principles from BitNet b1.58 \cite{wang2023bitnet}, which quantizes both weights and activations.
However, our approach departs from this; we systematically investigate three selective quantization strategies, visualized in the right side of Fig.~\ref{fig:method}, to identify the optimal balance between prediction accuracy and computational efficiency.
These strategies are: (a) \textbf{BitLinear-Both}, quantizing both weights and activations; (b) \textbf{BitLinear-Activ}, applying activation-only quantization; and (c) \textbf{BitLinear-Weight}, applying weight-only quantization.

\textbf{Scaling Functions and Quantization. }
The BitLinear operation is defined by its normalization and scaling procedures.
First, all input activations $x$ are normalized via LayerNorm ($\text{LN}$) to produce $x_{\text{norm}} = \text{LN}(x)$.
Second, two scaling factors, $\gamma$ for activations and $\beta$ for weights, are computed.
As visualized in Fig.~\ref{fig:method}-(a,b), for 8-bit activation quantization ($Q_{\text{activ}} = [-128, 127]$), we employ AbsMax scaling:
\begin{equation}\label{eq:absmax}
\begin{gathered}
\begin{aligned}
\gamma = \frac{\max(|Q_{\text{activ}}|)}{\max(|x_{\text{norm}}|) + \epsilon} = \frac{127}{\max(|x_{\text{norm}}|) + \epsilon}.
\end{aligned}
\end{gathered}
\end{equation}
For 1.58-bit ternary weight quantization ($Q_{\text{weight}} = \{-1, 0, 1\}$), visualized in Fig.~\ref{fig:method} (c), we utilize AbsMean scaling:
\begin{equation}\label{eq:absmean}
\begin{gathered}
\begin{aligned}
\beta = \frac{\max(|Q_{\text{weight}}|)}{\text{mean}(|W|) + \epsilon} = \frac{1}{\text{mean}(|W|) + \epsilon}.
\end{aligned}
\end{gathered}
\end{equation}
The core quantization function, denoted as $\text{Quant}_Q(z)$, then transforms the scaled values $z$ into the target integer range $Q$ using $\text{Round Clamp}$ operations.
A Straight-Through Estimator (STE) is employed during the backward pass to maintain gradient flow.

\textbf{Selective Quantization Strategies. }
Based on these functions, we define three BitLinear variants.
In each case, the final output $y$ is computed by performing the linear operation on the quantized values, followed by a de-quantization step (division by the relevant scaling factors) to restore the original scale.
\\\textbf{Bitlinear-Both. }
This variant mirrors the original BitNet~\cite{wang2023bitnet} by quantizing both weights and activations, targeting maximum theoretical efficiency:
\begin{equation}\label{eq:both}
\begin{gathered}
\begin{aligned}
y = \frac{\text{Quant}_{8}(\gamma \cdot \text{LN}(x)) \cdot \text{Quant}_{1.58}(\beta \cdot W)^T + b}{\beta \cdot \gamma}.
\end{aligned}
\end{gathered}
\end{equation}
\textbf{Bitlinear-Activ. }
This strategy maintains full-precision weights but quantizes the 8-bit activations after LayerNorm, aiming for computational savings during the matrix multiplication:
\begin{equation}\label{eq:activ}
\begin{gathered}
\begin{aligned}
y = \frac{\text{Quant}_{8}(\gamma \cdot \text{LN}(x)) \cdot W^T + b}{\gamma}.
\end{aligned}
\end{gathered}
\end{equation}
\textbf{Bitlinear-Weight. }
This strategy keeps activations in full precision while quantizing only the weights to 1.58-bit.
This strategy is designed to significantly reduce the model's memory footprint:
\begin{equation}\label{eq:weight}
\begin{gathered}
\begin{aligned}
y = \frac{x \cdot \text{Quant}_{1.58}(\beta \cdot W)^T + b}{\beta}.
\end{aligned}
\end{gathered}
\end{equation}

In the following sections, we empirically compare these three strategies and show that BitLinear-Weight achieves the best trade-off between accuracy and efficiency for LLM-based trajectory prediction.

%% file: sec/4_exps.tex
\section{Experiments}
\label{sec:experiments}

In this section, we present a comprehensive evaluation of our proposed BitTP framework.
We begin in Sec.~\ref{sec:setup} by detailing the experimental setup, including the widely-used ETH/UCY benchmark, our evaluation metrics, and key implementation details.
Next, in Sec.~\ref{sec:results}, we conduct a quantitative and qualitative comparison of our three BitTP strategies against state-of-the-art baselines, analyzing their predictive accuracy.
Following that, Sec.~\ref{sec:cost} provides a detailed analysis of the inference cost and memory usage for each configuration.
Finally, in Sec.~\ref{sec:lr}, we investigate training dynamics by analyzing the model's sensitivity to different learning rates, confirming the stability of our approach.
Furthermore, to demonstrate its suitability for edge devices and on-board computers, we validate the feasibility of deploying our model even in environments without GPU acceleration, which is detailed in Sec.~\ref{sec:cpu_eval} of the Supplementary Material.

\begin{figure*}[]
    \centering
    \includegraphics[width=1.0\linewidth]{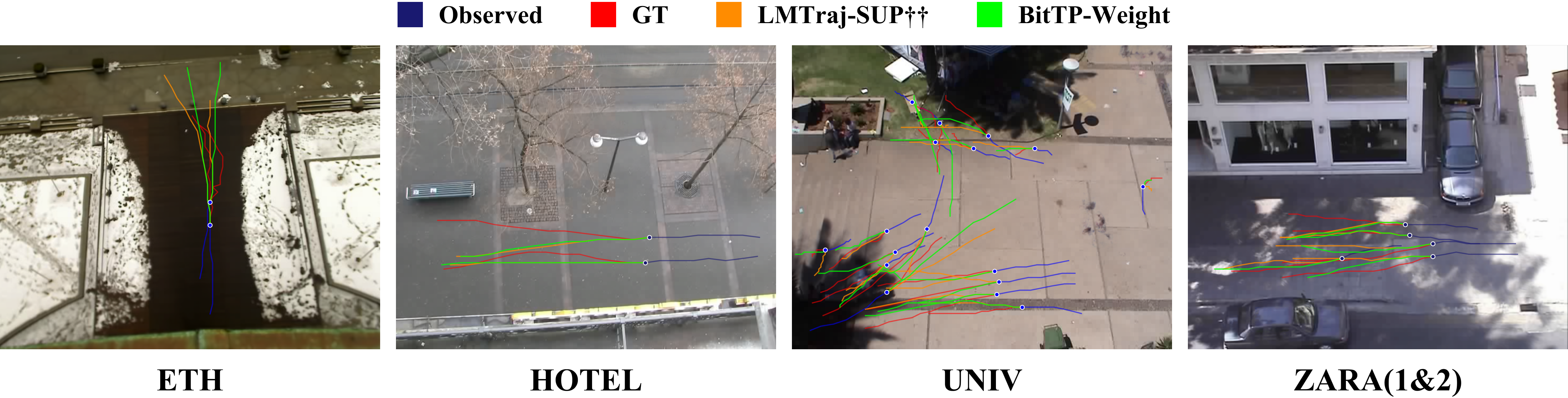}
    \caption{\textbf{Qualitative visualization of stochastic trajectory predictions on the ETH/UCY benchmark.}}
    \label{fig:qual}
\end{figure*}

\subsection{Experimental Setup}
\label{sec:setup}
\textbf{Datasets.}
We adopt the ETH/UCY trajectory datasets, which contain five scenes: \textit{ETH}, \textit{Hotel}, \textit{Univ}, \textit{Zara1}, and \textit{Zara2}.
Following the standard protocol, each sequence consists of 8 observed steps (3.2s) and 12 prediction steps (4.8s).
We employ a leave-one-out evaluation scheme, training on four scenes and testing on the remaining one.

\textbf{Metric.}
We report Average Displacement Error (ADE) and Final Displacement Error (FDE), where ADE is the mean positional error over the prediction horizon and FDE is the error at the final timestep.

\textbf{Model and Implementation Details.}
All experiments are conducted using the T5-small backbone (Fig.~\ref{fig:method}) implemented in PyTorch. 
BitTP models are trained on a single NVIDIA RTX 3090 GPU with AdamW optimizer and an initial learning rate of $1\times10^{-4}$ using a linear scheduler without warmup steps.
The training epoch is set to 8 with a batch size of 128 or 256.
Gradient clipping is applied with $\|g\|_{max}=1.0$.
We utilize a trajectory subword tokenizer trained via BPE \cite{Bae_2024_CVPR} on pixel-level trajectories to handle continuous input.
Stochastic decoding modes are evaluated with temperature 0.7.

\textbf{Quantization Settings.}
BitTP introduces lightweight \texttt{BitLinear} modules for weight-only, activation-only, and combined quantization, as described in Fig.~\ref{fig:method}-(a,b,c).
We evaluated multiple precision levels (1.58-bit, 4-bit, 8-bit) and compared them with a BF16 baseline \cite{Bae_2024_CVPR}.
During inference, stochastic decoding (temperature=0.7) is reported, with top-k sampling disabled.

\subsection{Performance Comparison Across BitTP Strategies}
\label{sec:results}
Tab.~\ref{tab:bitlinear_comparison} compares BitTP models with existing state-of-the-art stochastic trajectory prediction methods \cite{Bae_2024_CVPR,mangalam2020journeydestinationendpointconditioned,Gu_2022_CVPR,SocialVAE} on the ETH/UCY benchmark.
Methods such as SocialVAE and LMTraj-SUP achieve the strongest full-precision performance, with average ADE/FDE around 0.21/0.33 and 0.22/0.32, respectively.
These models represent the performance ceiling for encoder–decoder architectures on this benchmark and provide a solid reference point for evaluating quantized variants.

Before evaluating BitTP, we reproduced LMTraj-SUP \cite{Bae_2024_CVPR} under BF16, INT8, and INT4 settings.
The reproduced BF16 model (LMTraj-SUP††) shows slightly degraded performance compared to the original numbers (AVG 0.35/0.62 vs. 0.22/0.32).
Importantly, both INT8 (AVG 0.34/0.59, $\Delta$ -0.01/-0.03) and INT4 (AVG 0.34/0.60, $\Delta$ -0.01/+0.02) configurations maintain nearly identical accuracy, indicating that transformer-based trajectory models remain robust under moderate quantization~\cite{dettmers2022llmint8_neurips, dettmers2023qloraefficientfinetuningquantized, dettmers2022optimizers, dettmers2023case}.

The three BitTP strategies display distinctly different behaviors. BitTP-Weight (Fig.~\ref{fig:method}-(c)) achieves the strongest performance, with an average of 0.30 ADE and 0.49 FDE—improving upon the reproduced BF16 baseline ($\Delta$ -0.05/-0.13) and outperforming all quantized LMTraj-SUP settings.
This suggests that weight-only \texttt{BitLinear} layers effectively preserve spatial–temporal structure while benefiting from reduced parameter redundancy.
BitTP-Both (Fig.~\ref{fig:method}-(a)), however, shows degraded accuracy with an average of 1.27/1.13 ($\Delta$ +0.92/+0.51), as joint encoder–decoder quantization accumulates noise and disrupts attention-layer activations, consistent with the instability observed during training.
Lastly, BitTP-Activation (Fig.~\ref{fig:method}-(b)) performs the worst, achieving an average of 1.84/3.12 ($\Delta$ +1.49/+2.50), which reflects substantial instability when only activations are quantized; early activation distortion significantly increases gradient variance and disrupts feature-flow consistency.
BitTP-Weight (Fig.~\ref{fig:method}-(c)) achieves the strongest performance, with an average of 0.30 ADE and 0.49 FDE. This corresponds to a relative error reduction of 14.29\% in ADE and 20.97\% in FDE compared to the reproduced BF16 baseline ($\Delta$ -0.05/-0.13), outperforming all quantized LMTraj-SUP settings.

To complement our quantitative analysis, Fig.~\ref{fig:qual} presents a qualitative visualization of stochastic trajectory predictions from the ETH/UCY benchmark.
As shown in the scenes from ETH, HOTEL, UNIV, and ZARA, the trajectories predicted by our BitTP-Weight consistently generate plausible paths that closely align with the ground truth.
When compared to the full-precision LMTraj-SUP†† baseline, our lightweight model demonstrates a comparable or superior ability to capture complex multi-agent interactions and anticipate future movements.
This visual evidence supports our main finding that selective weight-only quantization effectively preserves the model's critical spatio-temporal reasoning capabilities.

Overall, BitTP-Weight provides the best accuracy and efficiency, demonstrating that low-bit trajectory transformers are feasible when quantization is applied selectively.
The weaker performance of BitTP-Both and BitTP-Activation highlights that stability in encoder–decoder architectures is highly sensitive to the location of quantization, emphasizing the importance of preferring weight quantization over activation quantization in practice.

\subsection{Inference Cost Comparison Across BitTP Strategies}
\label{sec:cost}
To understand the efficiency of different quantization designs, we compare the inference cost of each BitTP strategy using BitTP-Both as the reference baseline (100\% memory and 100\% inference cost) in Tab.~\ref{tab:avg_accu}.
BitTP-Both jointly quantizes all \texttt{nn.Linear} layers in both the encoder and decoder, and therefore represents the full quantization path without any structural reduction.
The relative cost of the other variants can thus be interpreted as the proportion of computation and memory they eliminate with respect to this baseline.

BitTP-Activation applies 8-bit quantization only to activations while keeping all weights in high precision.
Since most computation in transformer architectures arises from weight matrices, activation-only quantization reduces the overall load only marginally, resulting in 91.09\% memory and 74.68\% inference cost relative to BitTP-Both.
Importantly, this limited efficiency gain comes at the expense of a severe accuracy degradation (1.84 ADE / 3.12 FDE), as quantizing activations directly distorts intermediate feature distributions and destabilizes attention propagation.

BitTP-Weight, which quantizes all weight matrices to 1.58-bit while preserving full-precision activations, yields substantial efficiency improvements. Weight-only quantization reduces parameter storage and matrix multiplication cost, achieving 53.58\% memory and 63.12\% inference cost. Despite this reduction, BitTP-Weight maintains strong predictive performance (0.30 ADE / 0.49 FDE), surpassing all other quantized variants and closely matching the reproduced BF16 results.

Overall, these results demonstrate that BitTP-Weight is the most practical and efficient design, offering significant computational savings without compromising accuracy. In contrast, BitTP-Activation provides minimal benefit while severely damaging model stability, underscoring the importance of carefully selecting quantization locations in encoder–decoder architectures.

\begin{table}[t]
\centering
\caption{\textbf{Accuracy and complexity comparison between quantized and baseline models in ETH/UCY.}}
\label{tab:avg_accu}
\begin{tabular}{lcccc}
\toprule
 & \multicolumn{2}{c}{\textbf{Accuracy}} & \multicolumn{2}{c}{\textbf{Complexity(\%)}} \\
\cmidrule(lr){2-3} \cmidrule(lr){4-5}
\textbf{Model} & \textbf{ADE} & \textbf{FDE} & \textbf{Memory} & \textbf{Inference} \\
\midrule
BitTP-Both & 1.27 & 1.13 & 100.00 & 100.00 \\
BitTP-Activation & 1.84 & 3.12 & 91.09 & 74.68 \\
BitTP-Weight & 0.30 & 0.49 & 53.58 & 63.12 \\
\bottomrule
\end{tabular}
\end{table}

\begin{table}[t]
\centering
\normalsize
\caption{\textbf{Accuracy comparison across quantization and BitLinear configurations.} (batch\_size=256).}
\label{tab:accuracy_bitlinear}
\begin{tabular}{lcc}
\toprule
\textbf{Model} & \textbf{ADE} & \textbf{FDE} \\
\midrule
\midrule
\multicolumn{3}{l}{\textbf{Learning rate: 1e-4}} \\
LMTraj-SUP\cite{Bae_2024_CVPR}†† & 0.47 & 0.67 \\
BitTP-Both & 1.55 & 1.34 \\
BitTP-Activation & 2.89 & 4.86 \\
BitTP-Weight (Best) & \textbf{0.42} & \textbf{0.57} \\
\midrule
\multicolumn{3}{l}{\textbf{Learning rate: 2e-4}} \\
LMTraj-SUP\cite{Bae_2024_CVPR}†† & 0.54 & 0.81 \\
BitTP-Both & 1.71 & 1.67 \\
BitTP-Activation & 1.59 & 1.53 \\
BitTP-Weight & 0.51 & 0.79 \\
\midrule
\multicolumn{3}{l}{\textbf{Learning rate: 4e-4}} \\
LMTraj-SUP\cite{Bae_2024_CVPR}†† & 0.62 & 0.96 \\
BitTP-Both & 1.68 & 1.36 \\
BitTP-Activation & 1.36 & 1.23 \\
BitTP-Weight & 0.52 & 0.75 \\
\bottomrule
\end{tabular}
\end{table}

\subsection{Learning-Rate Effects in Encoder-Decoder}
\label{sec:lr}
Recent findings on BitNet and other decoder-only low-bit models show that unusually high learning rates can \textit{improve} optimization stability under aggressive quantization, smoothing the loss landscape and mitigating quantization-induced noise~\cite{wang2023bitnet}.
This behavior contrasts with conventional quantization pipelines—where higher learning rates typically destabilize training—and suggests that weight-only quantization may follow a distinct optimization regime.
However, prior analyses have focused almost exclusively on decoder-only LLMs. Whether this phenomenon extends to encoder–decoder architectures, where both sides jointly influence generation, has remained unexamined.

Tab.~\ref{tab:accuracy_bitlinear} addresses this gap. Across all learning rates (1e-4, 2e-4, 4e-4), \textbf{BitTP-Weight} remains remarkably stable and consistently outperforms the BF16 baseline, achieving its best accuracy at 1e-4 (0.42 ADE / 0.57 FDE).
These results indicate that the high–learning-rate robustness reported in BitNet generalizes naturally to seq2seq transformers: weight-only quantization preserves a smooth and optimization-friendly loss landscape even when both encoder and decoder participate in prediction.

\begin{figure}[h]
    \centering
    \includegraphics[width=0.9\linewidth]{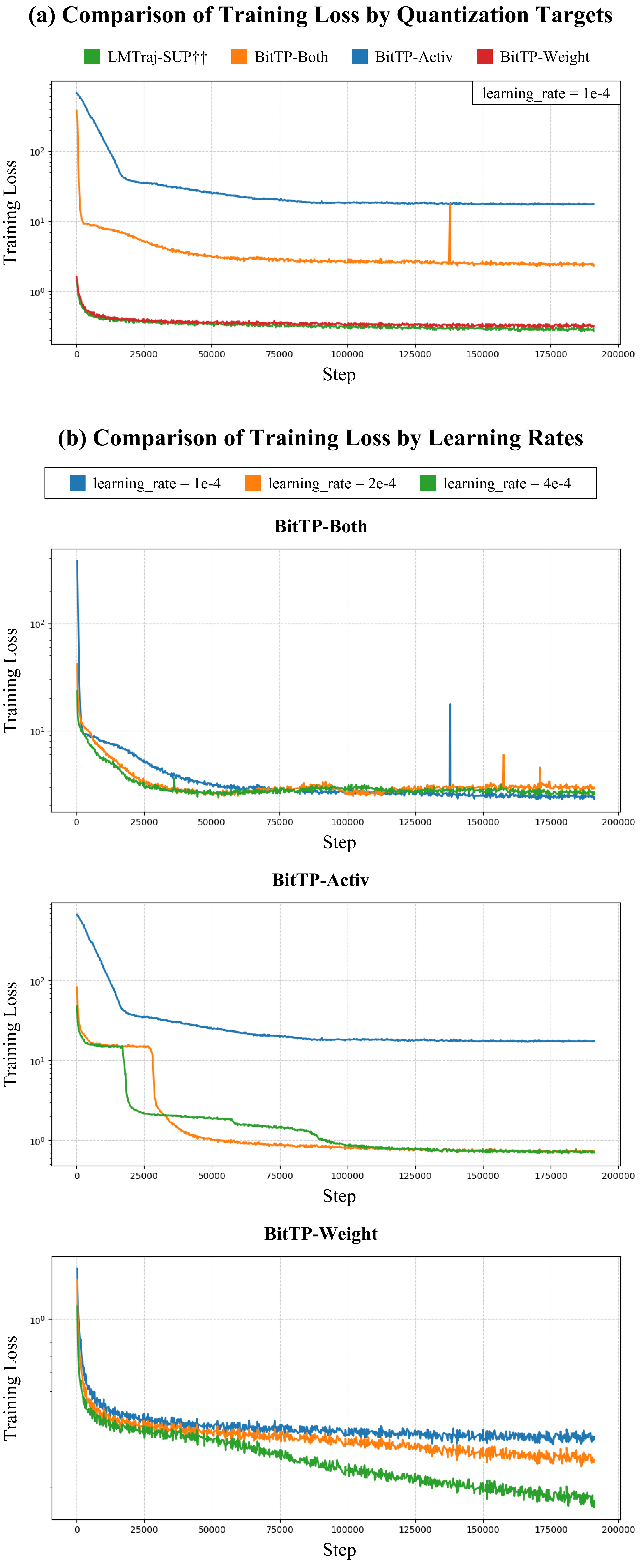}
    \caption{\textbf{Training loss convergence curves in ETH (Y-axis in log-scale).} (a) Comparison of quantization targets ($\text{lr}=1\times10^{-4}$). (b) Sensitivity analysis of different learning rates for each BitTP strategy (batch\_size=256).}
    \label{fig:loss}
    \vspace{-4mm}
\end{figure}

Fig.~\ref{fig:loss}-(a) provides deeper insight into these trends under a fixed learning rate (1e-4).
\textit{BitTP-Weight} converges smoothly and monotonically, nearly mirroring the BF16 (LMTraj-SUP) baseline. This confirms that reducing weight precision does not distort early optimization dynamics or impede gradient flow. In contrast, \textit{BitTP-Activation} exhibits a step-like training pattern: the loss remains flat for an extended period before abruptly dropping once certain optimization thresholds are crossed.
This suggests that activation-only quantization introduces early optimization barriers and may require a warm-up phase before meaningful learning can occur. This behavior aligns with the poor accuracy of BitTP-Activation in Tab.~\ref{tab:accuracy_bitlinear}. 
\textit{BitTP-Both} lies between these two extremes. Its convergence is slower and noisier than BitTP-Weight, yet far more stable than BitTP-Activation.
Occasional sharp loss spikes indicate that jointly quantizing both encoder and decoder introduces intermittent instability, likely due to accumulated quantization noise propagating across layers.

Fig.~\ref{fig:loss}-(b) examines how each strategy responds to different learning rates (1e-4, 2e-4, 4e-4).
\textit{BitTP-Weight} remains stable across all settings, with nearly identical convergence trajectories. \textit{BitTP-Both} also maintains reasonable stability, showing only minor fluctuations even as the learning rate increases.

These findings suggest that bit-linear quantization does not fundamentally impair optimization in encoder–decoder transformers; rather, it can even benefit from moderately increased step sizes, echoing observations from decoder-only BitNet models. Activation-only quantization again stands as the outlier. For \textit{BitTP-Activation}, higher learning rates magnify instability, producing sharp spikes and abrupt loss drops—showing that perturbations in activation precision are far more disruptive than low-bit weights.

In summary, Weight-only quantization learns reliably in encoder–decoder architectures, mirroring the favorable optimization behavior previously reported in decoder-only models.
Joint weight–activation quantization (\textit{BitTP-Both}) remains usable but introduces intermittent noise, while activation-only quantization (\textit{BitTP-Activation}) suffers from inherent optimization instability that limits its practical utility.

%% file: sec/5_conclusion.tex
\section{Conclusion}

In this paper, we addressed the challenge of deploying large-scale reasoning models for trajectory prediction on resource-constrained edge devices.
We proposed BitTP, a quantization framework that transforms a sequence-to-sequence Transformer into a lightweight bitlinear model, and systematically analyzed three selective quantization strategies: BitLinear-Both, BitLinear-Activation, and BitLinear-Weight.
Notably, BitTP-Weight resolves the severe training instability and degraded spatio-temporal reasoning seen in activation-quantized variants (BitTP-Both and BitTP-Activation), ultimately surpassing the full-precision baseline.
We hypothesize that weight-only quantization acts as an implicit regularizer, alleviating overfitting in the full-precision model—an effect that may be particularly pronounced in datasets with low trajectory diversity and dominant paths such as ETH.
In contrast, activation-involving strategies (BitTP-Both and BitTP-Activ) suffered from severe training instability and performance degradation, suggesting that preserving activation precision is critical for stable spatio-temporal reasoning in encoder-decoder frameworks.
In conclusion, our work validates that the BitTP-Weight approach makes it practically feasible to deploy sophisticated, LLM-based reasoning for real-time trajectory prediction on real-world edge devices, such as autonomous vehicles and mobile robots.

%% file: sec/6_acknowledgement.tex
\section*{Acknowledgements}

This work was supported by the National Research Foundation of Korea (NRF) grant funded by the Korea government (MSIT) (No. RS-2025-22802992), the Basic Science Research Program through the National Research Foundation of Korea (NRF) funded by the Ministry of Education (No. RS-2025-25420118), the Institute of Information \& Communications Technology Planning \& Evaluation (IITP) grants funded by the Korea government (MSIT) (No. RS-2025-25442149, LG AI STAR Talent Development Program for Leading Large-Scale Generative AI Models in the Physical AI Domain / No. RS-2025-02219277, AI Star Fellowship Support (DGIST)), and the InnoCORE program of the Ministry of Science and ICT (No. N10260003 / No. 26-InnoCORE-01).

%% file: sec/X_suppl.tex
\clearpage
\setcounter{page}{1}
\maketitlesupplementary
\section{Implementation Details and Evaluation Metrics}
\label{sup:impl}

This section provides a comprehensive breakdown of the model architecture integration, the specific metrics used for performance assessment, and the hardware protocols employed for efficiency benchmarking.

\subsection{Backbone and BitLinear Integration}
We expand on the implementation details of the BitTP architecture described in the main paper. Our backbone utilizes the \textbf{T5-small} encoder-decoder structure, consisting of 6 encoder blocks and 6 decoder blocks ($d_{model}=512, d_{ff}=2048, \text{heads}=8$).

\textbf{Quantization Injection Points.}
We apply our recursive replacement algorithm to inject \texttt{BitLinear} modules into specific sub-layers of the Transformer blocks. The target modules include:
\begin{itemize}
    \item \textbf{Attention Projections:} The Query ($\mathbf{W}_Q$), Key ($\mathbf{W}_K$), Value ($\mathbf{W}_V$), and Output ($\mathbf{W}_O$) matrices in both Self-Attention and Cross-Attention mechanisms.
    \item \textbf{Feed-Forward Networks (FFN):} The expansion ($\mathbf{W}_{i}$) and projection ($\mathbf{W}_{o}$) layers within the position-wise FFNs.
\end{itemize}

\subsection{Latency and Memory Measurement Protocol}
To ensure fair and reproducible comparisons across different quantization strategies, we conducted all efficiency measurements under a standardized environment that reflects our experimental configuration.

\begin{itemize}
    \item \textbf{Computing Device:} All benchmarks were performed on a single \textbf{NVIDIA RTX 3090} GPU with $24\text{GB}$ of VRAM.
    
    \item \textbf{Batch Size:} We set the inference batch size to $B=1$
    
    \item \textbf{Decoding Settings:} 
    Unless stated otherwise, measurements include the autoregressive generation time for the full prediction horizon $T_{\text{pred}}=12$. 
    Crucially, we employed \textbf{stochastic sampling} with a temperature parameter $\tau=0.7$ (rather than greedy decoding) to account for the computational overhead of the random sampling process required for generating diverse trajectories.
    
    \item \textbf{Precision \& Cache:} Computations were performed in mixed precision (BF16/FP32) with the Key-Value (KV) cache enabled to optimize sequential decoding throughput.
\end{itemize}
\subsection{Evaluation Metrics}
We evaluate the predictive performance using standard trajectory prediction metrics. Since our model generates multimodal predictions via stochastic sampling, we report the Best-of-$K$ metrics, where the sample closest to the ground truth is selected for evaluation. In all experiments, we use $K = 20$ stochastic samples per prediction. 1000 trajectory samples are generated for each input in the main experiment, while only 1 sample per input is used in the CPU-based experiments in Sec.~\ref{sec:cpu_eval}.

Let $\mathbf{Y}_i \in \mathbb{R}^{T \times 2}$ be the ground truth trajectory for pedestrian $i$, and $\hat{\mathbf{Y}}_{i}^{(k)}$ be the $k$-th predicted sample among $K$ generated trajectories. If the dataset provides homography matrices $H$, all coordinates are projected from pixel space to world coordinates (meters) via a mapping function $\mathcal{M}(\cdot, H)$ before error calculation.

\begin{itemize}
    \item \textbf{Average Displacement Error (ADE):} 
    The mean Euclidean distance between the predicted path and the ground truth over all time steps. We report the minimum ADE (minADE) across $K$ samples, averaged over all $N$ pedestrians in the dataset:
    \begin{equation}
        \text{ADE} = \frac{1}{N} \sum_{i=1}^{N} \min_{k=1}^{K} \left( \frac{1}{T} \sum_{t=1}^{T} \lVert \hat{\mathbf{y}}_{t, i}^{(k)} - \mathbf{y}_{t, i} \rVert_2 \right)
    \end{equation}

    \item \textbf{Final Displacement Error (FDE):} 
    The Euclidean distance between the predicted position and the ground truth at the final time step $T$. Similarly, we report the minimum FDE (minFDE):
    \begin{equation}
        \text{FDE} = \frac{1}{N} \sum_{i=1}^{N} \min_{k=1}^{K} \left( \lVert \hat{\mathbf{y}}_{T, i}^{(k)} - \mathbf{y}_{T, i} \rVert_2 \right)
    \end{equation}

    \item \textbf{Peak Memory:} The peak VRAM usage required to load the model parameters, measured in Megabytes (MB). This metric directly reflects the compression ratio achieved by quantization.
    
    \item \textbf{Inference Latency:} The average wall-clock time (milliseconds) required to generate a full trajectory ($T=12$) for a single agent ($B=1$).
\end{itemize}

\begin{table*}[t]
    \centering
    \small
    \renewcommand{\arraystretch}{1.2}
    \caption{\textbf{Wall-Clock Training Time Comparison across All Datasets.} Measured on a single NVIDIA RTX 3090 with a fixed batch size of 128 and learning rate of $1\times 10^{-4}$ for 8 epochs. The symbol $\times$ denotes the relative time increase compared to the baseline.}
    \label{tab:full_training_time}
    \begin{tabular}{l|c|ccccc|c}
    \toprule
    \multirow{2}{*}{\textbf{Model Variant}} & \multirow{2}{*}{\textbf{Target}} & \multicolumn{5}{c|}{\textbf{Training Duration (Hours)}} & \multirow{2}{*}{\textbf{Avg. Overhead}} \\
    \cmidrule(lr){3-7}
     & & \textbf{ETH} & \textbf{Hotel} & \textbf{Univ} & \textbf{Zara1} & \textbf{Zara2} & \\
    \midrule
    LMTraj-SUP (Baseline) & BF16 & 40h & 33h & 9h & 31h & 28h & 1.00$\times$ \\
    \midrule
    BitTP-Both & Both & 72h & 60h & 18h & 56h & 52h & 1.90$\times$ \\
    BitTP-Activ & Activations only & 68h & 59h & 18h & 55h & 52h & 1.85$\times$ \\
    BitTP-Weight & Weights only & 46h & 40h & 12h & 37h & 35h & 1.20$\times$ \\
    \bottomrule
    \end{tabular}
\end{table*}

\subsection{Latency and Memory Measurement Protocol}
To ensure fair and reproducible comparisons across different quantization strategies, we conducted all efficiency measurements under a standardized environment that reflects our experimental configuration.

\begin{itemize}
    \item \textbf{Computing Device:} All benchmarks were performed on a single \textbf{NVIDIA RTX 3090 (24GB)} GPU.
    \item \textbf{Batch Size:} We utilized a batch size of $1$, online trajectory prediction scenario
    \item \textbf{Decoding Settings:} 
    Unless stated otherwise, measurements include the autoregressive generation time for the full prediction horizon ($T_{pred}=12$). 
    Crucially, we employed \textbf{stochastic sampling} with a temperature of $\tau=0.7$ (rather than greedy decoding) to account for the computational overhead of the random sampling process required for generating diverse trajectories.
    \item \textbf{Precision \& Cache:} Computations were performed in mixed precision (BF16/FP32) with the Key-Value (KV) cache enabled to optimize sequential decoding throughput.
\end{itemize}

\section{Quantization Methods: Existing Techniques}
\label{sup:quant_methods}

This section elaborates on the theoretical underpinnings and implementation details of the quantization strategies compared in this study. For the post-training quantization (PTQ) baselines, we leverage the optimized kernels provided by the \texttt{bitsandbytes} library~\cite{dettmers2022llmint8_neurips, dettmers2023qloraefficientfinetuningquantized}, which implement vector-wise quantization and normal-float data types. We contrast these with our proposed BitTP, which fundamentally alters the linear projection mechanics.

\subsection{INT8 Baseline: Vector-wise Quantization with Decomposition}
For the 8-bit baseline, we employ the \textbf{LLM.int8()} methodology. Unlike standard abs-max quantization which can be degraded by outlier features, this method utilizes a \textit{mixed-precision decomposition} strategy.

\textbf{Vector-wise Quantization.} 
The matrix multiplication $\mathbf{Y} = \mathbf{X}\mathbf{W}$ is approximated by scaling rows of $\mathbf{X}$ and columns of $\mathbf{W}$. Let $C_x$ and $C_w$ be the scaling constants for the input and weight tensors, respectively. The quantized computation is defined as:
\begin{equation}
    \mathbf{Y} \approx \text{dequant}\left( \text{quant}(\mathbf{X}) \cdot \text{quant}(\mathbf{W}) \right) = S_x \cdot \mathbf{X}_{\text{int8}} \cdot \mathbf{W}_{\text{int8}} \cdot S_w^T
\end{equation}
where $S_x$ and $S_w$ preserve the dynamic range of the activations and weights.

\textbf{Outlier Handling (Decomposition).} 
To handle emergence of extreme outliers in Transformer activations, the matrix multiplication is decomposed into two parts based on an outlier threshold $\alpha$:
\begin{equation}
    \mathbf{Y} = \underbrace{(\mathbf{X}_{out} \cdot \mathbf{W}_{out})}_{\text{BF16}} + \underbrace{(\mathbf{X}_{int} \cdot \mathbf{W}_{int})}_{\text{INT8}}
\end{equation}
Dimensions where $|x_i| > \alpha$ are extracted and computed in high precision (BF16), while the bulk of dimensions are computed in INT8. This ensures numerical stability but introduces the inference latency overhead observed in our experiments.

\subsection{4-bit Baseline: NormalFloat (NF4) Quantization}
For the 4-bit baseline, we adopt the \textbf{NormalFloat4 (NF4)} data type. This approach assumes that the weights of a converged neural network follow a zero-mean normal distribution $\mathcal{N}(0, 1)$ and employs \textit{Quantile Quantization}.

\textbf{Information-Theoretic Optimality.} 
Standard integer quantization relies on linear spacing, which is suboptimal for bell-curved distributions. NF4 constructs quantization levels $q_i$ such that each bin has equal probability mass. The levels are determined by the quantiles of the normal distribution:
\begin{equation}
    q_i = Q^{-1}\left(\frac{i}{2^k + 1}\right)
\end{equation}
where $Q^{-1}$ is the inverse cumulative distribution function (CDF) of $\mathcal{N}(0, 1)$ and $k=4$ bits. During the forward pass, these 4-bit stored weights are dequantized on-the-fly to BF16 for computation, prioritizing memory efficiency over arithmetic speedup.

\section{Training-Time Behavior}
\label{sup:training}

\begin{figure*}[t]
    \centering
    \includegraphics[width=1.0\linewidth]{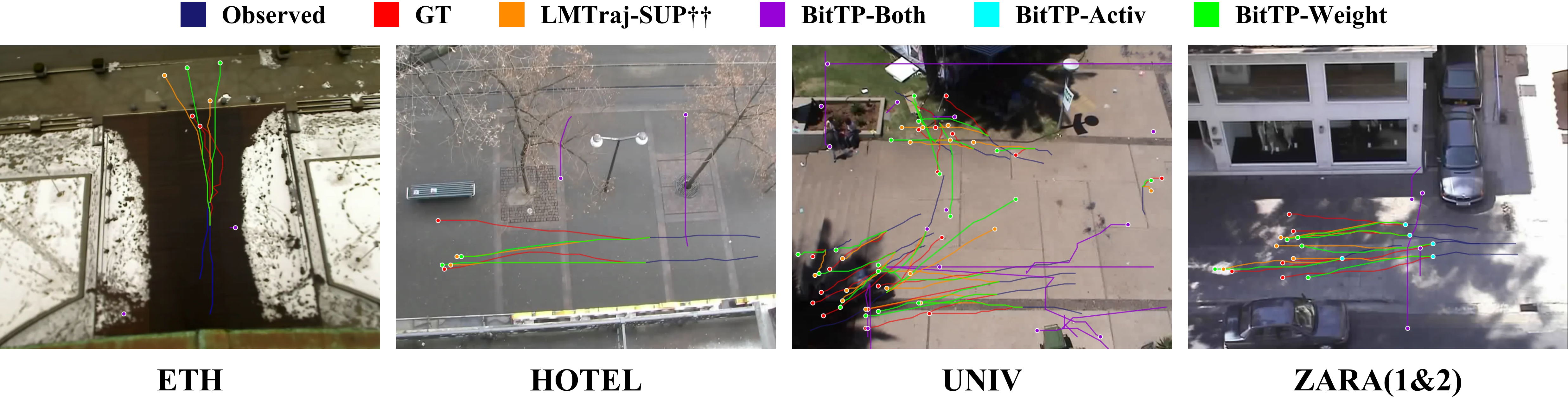}
    \caption{\textbf{Qualitative comparison of predicted pedestrian trajectories on the ETH/UCY scenes (ETH, HOTEL, UNIV, ZARA1\&2) under different quantization strategies.} Observed past trajectories, ground-truth futures, and predictions from LMTraj-SUP††, BitTP-Both, BitTP-Activ, and BitTP-Weight are overlaid; BitTP-Activ frequently collapses and fails to produce plausible future trajectories, while BitTP-Weight preserves accurate trajectory shapes.}
    \label{fig:qual_all}
\end{figure*}

In this section, we analyze the optimization dynamics of training BitTP variants from scratch (or fine-tuning) compared to the BF16 baseline. We focus on training stability and computational overhead. While the primary goal of BitTP is to reduce inference memory and latency, analyzing training efficiency is crucial for understanding the cost of using \texttt{BitLinear}.

\textbf{Measurement Setup.}
We measured the wall-clock training time on a single \textbf{NVIDIA RTX 3090 (24GB)} GPU. The training configuration was standardized with a batch size of 128 and a learning rate of $1\times 10^{-4}$ over 8 epochs.

\textbf{Training Duration and Overhead.}
Tab.~\ref{tab:full_training_time} presents the total training duration required across five datasets (ETH, Hotel, Univ, Zara1, Zara2). We observe distinct overhead patterns depending on the quantization target:

\begin{itemize}
    \item \textbf{Weight-Only Efficiency:} The \textit{BitTP-Weight} variant incurs a relatively minor overhead (avg. $\sim 20\%$) compared to the baseline. This is because weight quantization (ternarization) involves element-wise operations that are computationally inexpensive relative to the full matrix multiplication cost.
    
    \item \textbf{Activation Quantization Cost:} The \textit{BitTP-Both} and \textit{BitTP-Activ} variants show a significant increase in training time ($\sim 1.8\text{--}1.9\times$). This substantial overhead is attributed to the \textbf{dynamic activation quantization} process. Unlike weights, activations change every forward pass, requiring on-the-fly calculation of scaling factors (e.g., AbsMean) and rounding operations for every token. In our Python-based implementation, this creates a computational bottleneck that limits training throughput.
\end{itemize}
\textbf{Consistency Across Datasets.}
This trend was consistent across all datasets. For instance, in the Univ dataset, training BitTP-Weight required 12 hours, whereas BitTP-Both required 18 hours ($1.5\times$), reinforcing the observation that the computational bottleneck lies primarily in the activation quantization path.

\section{Qualitative Results of Forecasting the Trajectory}
\label{sup:qualitative}

\begin{figure*}[h]
    \centering
    \includegraphics[width=1.0\linewidth]{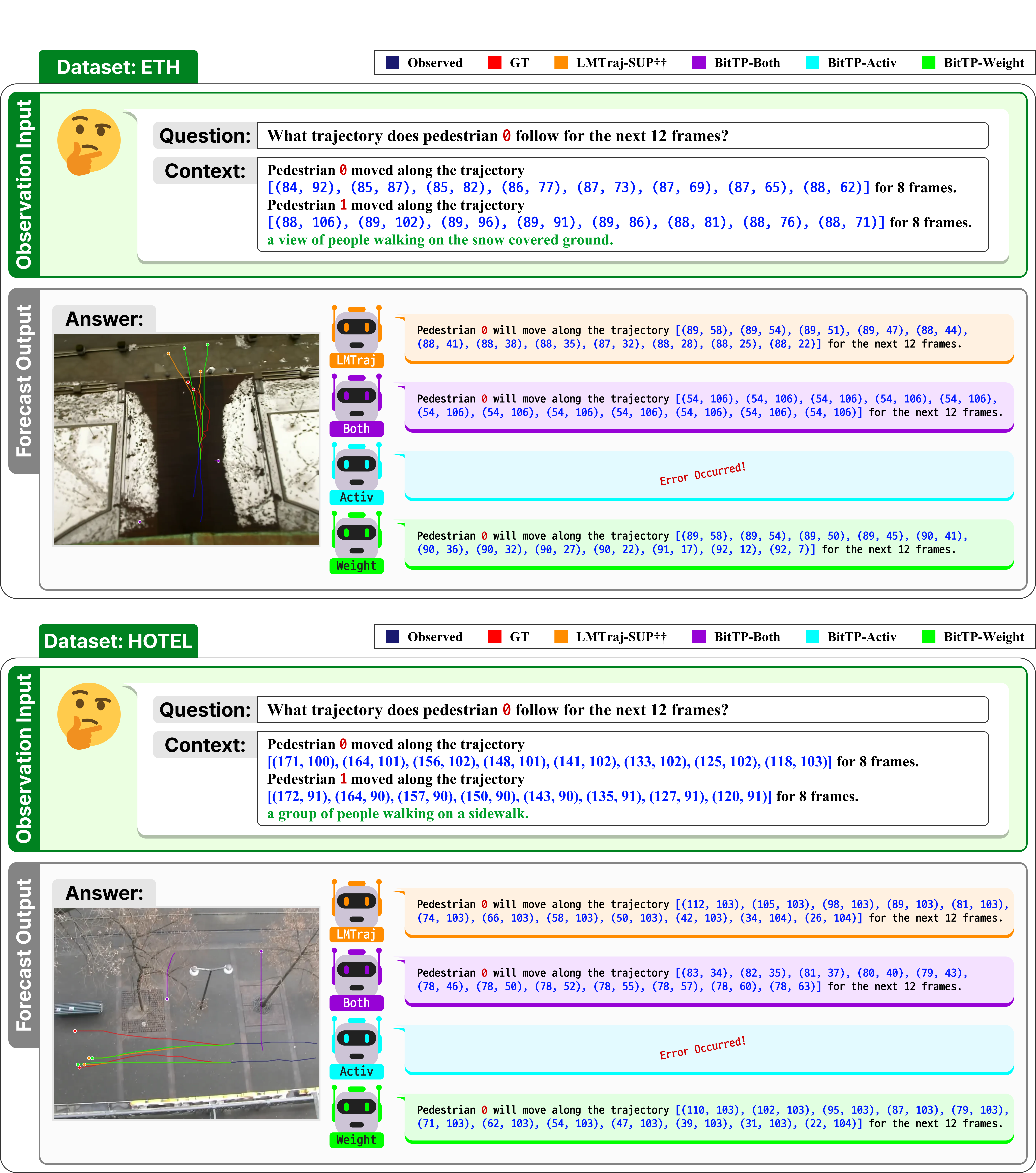}
    \caption{\textbf{Qualitative comparison of predicted trajectories on the ETH and HOTEL scenes under different quantization strategies for the trajectory forecast task.} For each example, the input observation consists of a question and a context. Only outputs for pedestrian 0 are visualized. Red denotes the pedestrian number, blue denotes the predicted trajectory, and green denotes the context; all colored elements are placeholders that can be replaced.}
    \label{fig:qual_eth_hotel}
\end{figure*}

\begin{figure*}[h]
    \centering
    \includegraphics[width=1.0\linewidth]{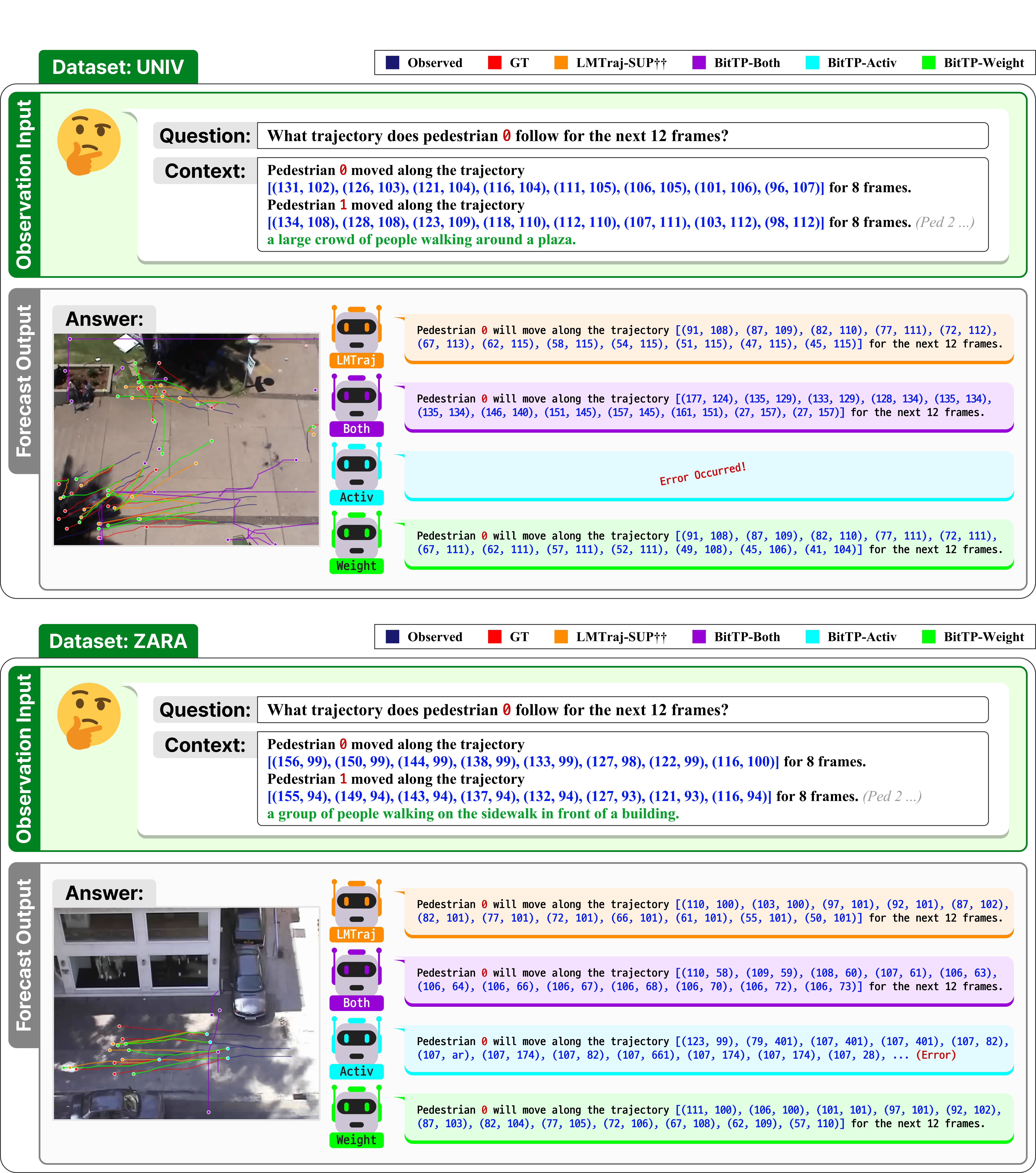}
    \caption{\textbf{Qualitative comparison of predicted trajectories on the UNIV and ZARA scenes under different quantization strategies for the trajectory forecast task.} For each example, the input observation consists of a question and a context. Only outputs for pedestrian 0 are visualized. Red denotes the pedestrian number, blue denotes the predicted trajectory, and green denotes the context; all colored elements are placeholders that can be replaced.}
    \label{fig:qual_univ_zara}
\end{figure*}

This section provides qualitative analyses of language-conditioned trajectory forecasting under different quantization strategies.
Fig.~\ref{fig:qual_all}-\ref{fig:qual_univ_zara} visualize stochastic predictions on the ETH/UCY benchmark and compare the full-precision LMTraj-SUP$\dagger\dagger$ model with BitTP-Weight, BitTP-Both, and BitTP-Activ.
Together, these figures illustrate how weight-only quantization preserves scene-consistent motion, whereas activation quantization leads to severe degradation in both trajectory generation and language outputs, complementing the quantitative results in the main paper.

\textbf{Multi-scene comparison.}  
Fig.~\ref{fig:qual_all} summarizes qualitative trajectory forecasts across five scenes (ETH, HOTEL, UNIV, ZARA1\&2) under different quantization strategies.
In all scenes, LMTraj-SUP$\dagger\dagger$ and BitTP-Weight produce future trajectories that closely align with the ground-truth (GT).
In contrast, BitTP-Both and BitTP-Activ fail to match the GT in a much more drastic manner.
BitTP-Both still generates trajectories; however, the predicted paths are often clearly unrealistic and significantly displaced from the GT routes, sometimes bending into unrealistic directions or overshooting the intended goal.
BitTP-Activ exhibits the most severe failure modes: for many examples, it does not produce a usable trajectory at all, either terminating without valid output or entering an effectively unbounded generation regime.
The ZARA example is particularly illustrative—although points are plotted, they stem from such runaway decoding that no coherent trajectory can be drawn.

\textbf{Input and output in the forecasting task.}  
Fig.~\ref{fig:qual_eth_hotel} and \ref{fig:qual_univ_zara} decompose the qualitative results from Fig.~\ref{fig:qual_all} by explicitly showing the language-based input and output for representative examples on the ETH/HOTEL and UNIV/ZARA scenes.
For each case, the observation is formatted as a question–context pair: the context encodes the past trajectory as a token sequence, and the question requests the future motion of pedestrian~0 over the forecasting horizon.
The figures then display both the generated text answer and the corresponding trajectory tokens rendered in the scene.
LMTraj-SUP$\dagger\dagger$ and BitTP-Weight produce well-structured answers that correctly refer to the requested forecast (e.g., horizon length, direction of movement) and generate coordinate tokens that form a coherent, scene-consistent trajectory when visualized.
In contrast, BitTP-Both and BitTP-Activ reveal how activation-involving quantization corrupts the decoding process itself: the outputs contain incomplete or repetitive phrases, or excessively long token streams that never settle into a valid trajectory sequence.
Fig.~\ref{fig:qual_eth_hotel} and \ref{fig:qual_univ_zara}, therefore, complement Fig.~\ref{fig:qual_all} by making explicit how the same failure patterns seen at the trajectory level already manifest at the language-interface level, while BitTP-Weight maintains reliable question–answer behavior and accurate future-path generation.

\begin{figure}[t]
    \centering
    \includegraphics[width=1.0\linewidth]{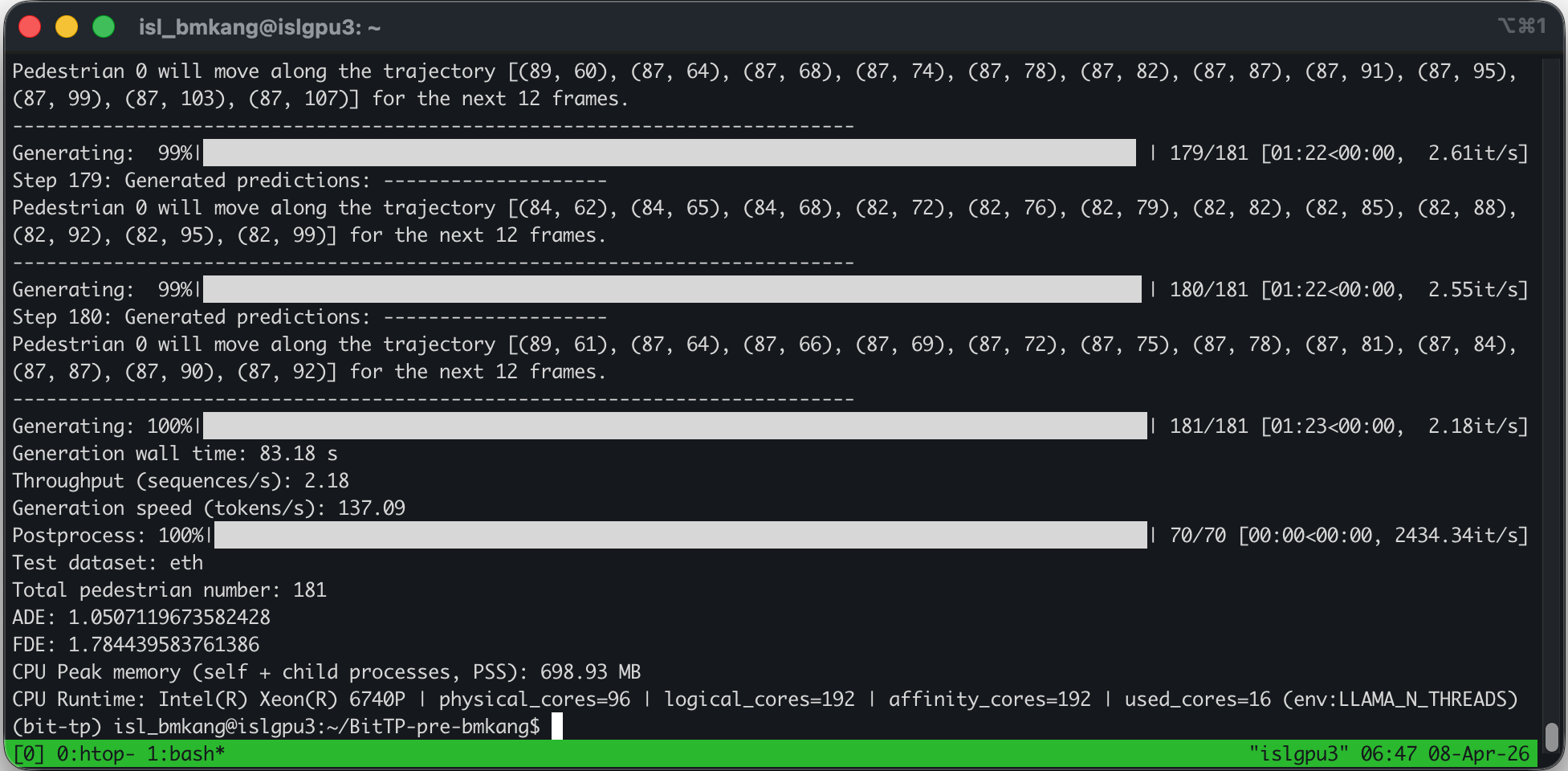}
    \caption{\textbf{Real-time execution log of BitTP-Weight in a CPU-only environment.} The screenshot displays the terminal output during inference on the ETH dataset using the \texttt{llama.cpp} framework, including runtime metrics and hardware utilization (Intel Xeon 6740P). Qualitative video demonstrations are available at: \url{https://mintcat98.github.io/BitTP/}}
    \label{fig:cpu_log}
\end{figure}

\section{CPU-based Inference and Cost Evaluation}
\label{sec:cpu_eval}
Fig.~\ref{fig:cpu_log} presents the execution log of BitTP in a CPU-only environment, exploring the feasibility of deploying an LLM-based predictor under strict hardware constraints. Despite the inherently massive computational requirements of LLMs, by leveraging Quantization-Aware Training (QAT) and a specialized \textbf{TQ1 kernel} \cite{vaidhya2025spectra} within the \texttt{llama.cpp} \cite{llamacpp} framework, BitTP-Weight successfully achieves a throughput of 2.18~sequences/s on the ETH subset. This result confirms that running sophisticated trajectory prediction models entirely on resource-constrained on-board computers without GPU acceleration is practically possible. 
To measure realistic edge-device latency, our CPU evaluation uses single-sample inference, departing from the standard 1,000-sample GPU protocol used for reproducible benchmarking. While this naturally introduces slight stochastic variance, it effectively demonstrates the model's practical execution speed under strict resource constraints.

Furthermore, the execution log directly validates our memory efficiency analysis: the peak CPU memory consumption (PSS) was measured at only 698.93~MB (including child processes), confirming the exceptionally lightweight memory footprint of our ternary-quantized model. Collectively, these results demonstrate that BitTP-Weight, when aggressively quantized and paired with a highly-optimized inference engine, significantly lowers the hardware barrier, opening up new possibilities for deploying LLMs on CPU-bound edge platforms.